\documentclass[lettersize,journal]{IEEEtran}
\usepackage{amsmath,amsfonts}
\usepackage{algorithmic}
\usepackage{array}
\usepackage[caption=false,font=normalsize,labelfont=sf,textfont=sf]{subfig}
\usepackage{textcomp}
\usepackage{stfloats}
\usepackage{url}
\usepackage{verbatim}
\usepackage{graphicx}
\usepackage{bm}

\usepackage{makecell}
\usepackage[table]{xcolor}

\usepackage{times}
\usepackage{microtype}
\usepackage{epsfig}
\usepackage[table,xcdraw]{xcolor}
\usepackage{caption}
\usepackage{float}
\usepackage{placeins}
\usepackage{color, colortbl}
\usepackage{stfloats}
\usepackage{enumitem}
\usepackage{tabularx}
\usepackage{xstring}
\usepackage{multirow}
\usepackage{multicol}
\usepackage{xspace}
\usepackage{url}
\usepackage{subcaption}
\usepackage{xcolor}
\usepackage{algorithm}
\usepackage{algorithmic}
\usepackage[hang,flushmargin]{footmisc}
\usepackage[normalem]{ulem}
\usepackage[pagebackref,breaklinks,colorlinks]{hyperref}
\usepackage{booktabs}
\usepackage{amsmath}
\usepackage{amssymb}
\usepackage{hyperref}  
\usepackage{enumitem}

\usepackage[dvipsnames]{xcolor}
\usepackage{rotating}
\definecolor{mygray}{gray}{0.8}

\begin{document}

\title{P2L-CA: An Effective Parameter Tuning \\ Framework for Rehearsal-Free Multi-Label Class-Incremental Learning}
\author{
Songlin Dong\IEEEauthorrefmark{1},
Jiangyang~Li,
Chenhao~Ding,
Zhiheng~Ma,
Haoyu~Luo,
Yuhang He,
Yihong Gong, \IEEEmembership{Fellow,~IEEE}
\\
\thanks{ Songlin Dong and Zhiheng Ma are with Faculty of Computility Microelectronics, Shenzhen University of Advanced Technology; Jiangyang~Li, Haoyu~Luo, Yihong Gong, and Yuhang He are with the College of Artificial Intelligence, Xi’an Jiaotong University; Chenhao Ding is with the College of Software Engineering, Xi’an Jiaotong University, Xi'an 710049, Shaanxi, China. (e-mail: \{dongsl, mazhiheng\}@suat-sz.edu.cn  \{dch225739, ljy1021, 3123358165\}@stu.xjtu.edu.cn; heyuhang@xjtu.edu.cn; ygong@mail.xjtu.edu.cn).}}




\markboth{Journal of \LaTeX\ Class Files,~Vol.~14, No.~8, August~2025}%
{Li \MakeLowercase{\textit{et al.}}: P2L-CA: An Effective Parameter Tuning \\ Framework for Multi-Label Class-Incremental Learning}

\maketitle

\begin{abstract}
Multi-label Class-Incremental Learning (MLCIL) aims to continuously recognize novel categories in complex scenes where multiple objects co-occur. However, existing approaches often incur high computational costs due to full-parameter fine-tuning and substantial storage overhead from memory buffers, or they struggle to address feature confusion and domain discrepancies adequately. To overcome these limitations, we introduce P2L-CA, a parameter-efficient framework that integrates a Prompt-to-Label (P2L) module with a Continuous Adapter (CA) module. The P2L module leverages class-specific prompts to disentangle multi-label representations while incorporating linguistic priors to enforce stable semantic–visual alignment. Meanwhile, the CA module employs lightweight adapters to mitigate domain gaps between pre-trained models and downstream tasks, thereby enhancing model plasticity. Extensive experiments across standard and challenging MLCIL settings on MS-COCO and PASCAL VOC show that P2L-CA not only achieves substantial improvements over state-of-the-art methods but also demonstrates strong generalization in CIL scenarios, all while requiring minimal trainable parameters and eliminating the need for memory buffers.
\end{abstract}


\section{Introduction}

Multi-label classification~\cite{asl2020,C-Trans,mlc_csvt1,mlc_csvt2}s has drawn remarkable research interest in the past decades and has wide applications in image search and medical image recognition. 
In recent years, with the rapid progress of social media and the constantly emerging new objects (such as the Cybertruck released lately), there is an increasing need to continuously and incrementally learn and recognize objects of unknown categories that are newly encountered, which is referred to as the multi-label class-incremental learning (MLCIL) problem~\cite{KRT,ML-SK}. This is more challenging than incremental image classification, as images may comprise multiple objects belonging to both old and new classes. In contrast, only the new classes are annotated in each training phase.

\begin{figure}[t]
\setlength{\abovecaptionskip}{-0 cm} 
\centering
\includegraphics[width=1\linewidth]{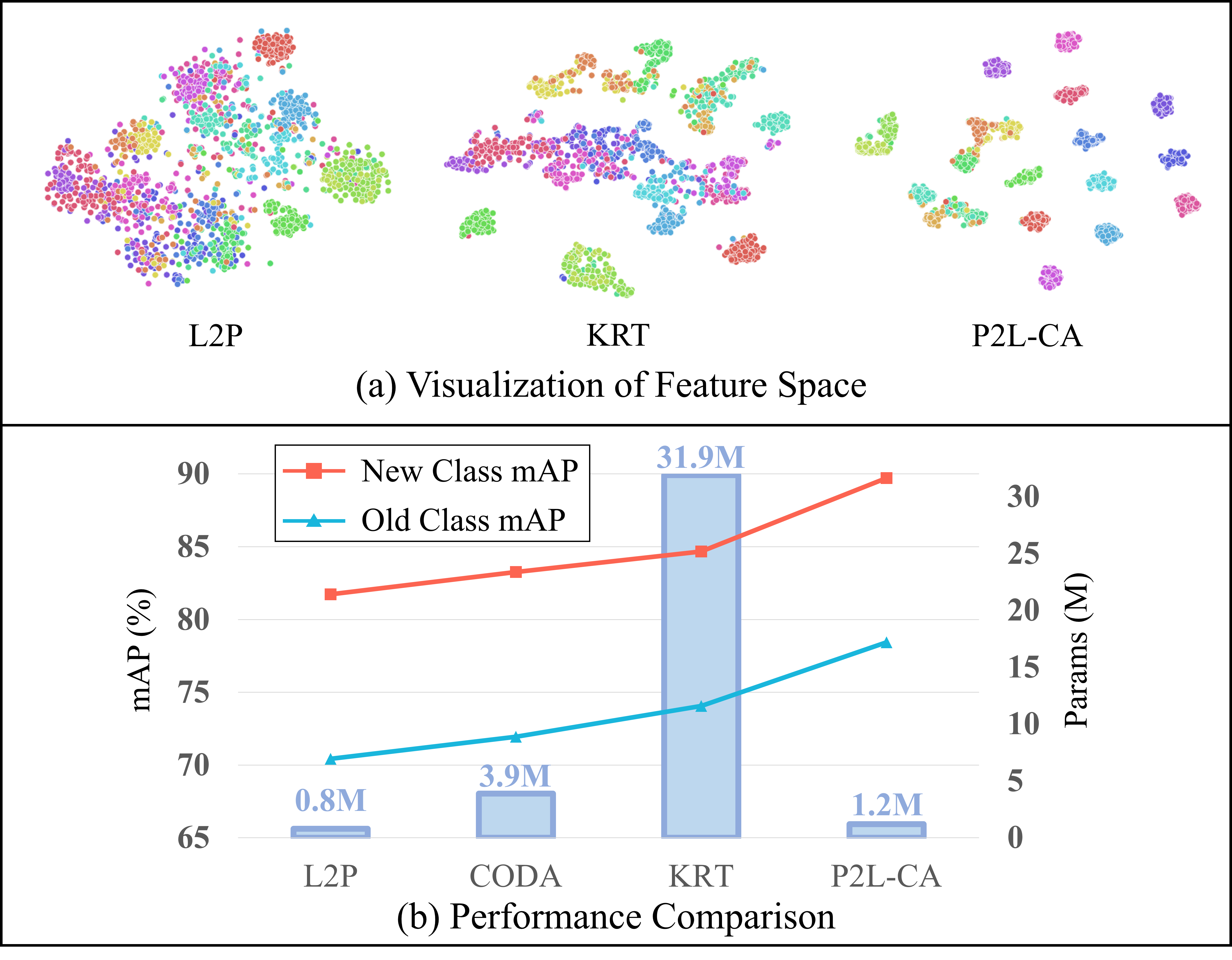}
\caption{\textbf{Effectiveness of the P2L-CA Framework}: (a) We present the t-SNE visualizations of 20 classes in COCO, illustrating that our method effectively addresses \textbf{feature confusion}. (b) We highlight that the CA module in P2L-CA effectively bridges the domain gap, leading to superior \textbf{transfer ability} and resulting in higher mAP scores for new classes.}
\label{Fig.mAP_base}
\end{figure}

To tackle this issue, existing MLCIL methods are primarily categorized into two paradigms: (i) Replay-based methods: Predominantly tailored for online learning scenarios, these approaches~\cite{PRS_mloil,OCDM,wang2025cut,cil_csvt7} rely on selecting and storing a subset of samples from previous tasks to approximate historical distributions. However, such buffer-based mechanisms not only suffer from strict storage constraints in standard settings but also pose severe privacy leakage risks when handling sensitive data. (ii) Model-based methods: These methods employ task-level attention modules~\cite{KRT}, task-specific GCNs~\cite{CSC2024}, or regularization techniques~\cite{rebll2025,cil_csvt6} to perform \textbf{full parameter fine-tuning} on the backbone network. However, this paradigm exhibits significant drawbacks: First, full parameter training incurs substantial computational and storage overheads (as illustrated in Fig.~\ref{Fig.mAP_base}); Second, the network parameters tend to be severely biased towards the current task, resulting in \textbf{feature confusion}; Furthermore, these methods often rely on pseudo-labeling techniques~\cite{KRT,rebll2025,ML-SK} to facilitate training, which necessitates additional forward inference steps, making efficient end-to-end learning difficult to achieve.

Recently, parameter-efficient fine-tuning (PEFT), particularly prompt-tuning~\cite{l2p,coda-prompt}, has achieved notable success in single-label IL. By freezing the backbone and introducing lightweight learnable prompts, these methods deliver strong anti-forgetting performance even in rehearsal-free settings. However, extending this paradigm to MLCIL exposes inherent limitations.
First, existing image-level~\cite{l2p} or task-level~\cite{dualprompt} prompts are overly coarse-grained and fail to disentangle the complex inter-class relationships in multi-label images, inevitably causing \textbf{feature confusion}.
Second, although simple prompt tuning is adequate for single-label tasks, its limited plasticity under a frozen backbone is insufficient for handling the substantial \textbf{domain gap} in multi-label scenarios, which strictly constrains the model’s feature transferability~(As shown in Fig.~\ref{Fig.mAP_base}).

This study proposes an efficient framework for the MLCIL task named \textit{P2L-CA}. Specifically, instead of directly modifying the model parameters, we introduce the \textit{prompt to label}~(P2L) module and the \textit{continuous adapter}~(CA) module to tackle the issues of feature confusion and domain gap, respectively.

Firstly, the P2L module aims to achieve incremental learning while separating different multi-label categories. Specifically, it consists of \textit{Class-Specific Prompts} and \textit{Feed Forward Network classifiers}. The former consists of a set of special prompts embedded into the encoder, characterized by continuous expansion across incremental stages, with each prompt corresponding to a specific class. In each incremental stage, these new trainable prompts and frozen prompts from previous tasks respectively facilitate learning of new classes and prevent forgetting of old classes. Simultaneously, the FFN classifier undergoes a similar incremental expansion, achieving a one-to-one correspondence between prompts and class labels. As shown in Fig.~\ref{Fig.mAP_base}~(a), the P2L module can extract label-level information end-to-end, achieving separation between classes and effectively preventing feature confusion. Then,
to further augment the representational power of these class-specific prompts, we introduce a semantic-aware initialization strategy leveraging a pre-trained language model~\cite{clip}. Instead of learning prompts from scratch, we extract semantic embeddings directly from category names. By projecting these rich semantic priors into the feature space, we explicitly model the semantic relationships between categories. This acts as a powerful guidance that assists the P2L module in learning more discriminative features, ensuring that the visual representations are semantically aligned with their corresponding class labels.

Secondly, we found that the low plasticity of the method stems from substantial domain discrepancies between the pre-training domain and downstream tasks. Therefore, we introduce the \textit{Continuous Adapter} module to leverage the strong transferability of adapters, reducing domain discrepancy and enhancing the plasticity of new classes. Concretely, we insert trainable adapters into a pre-trained ViT model and train the inserted adapters in the first stage to diminish domain differences and learn the label distribution of new domains. In the incremental stage, we freeze the CA module and only adjust classifier parameters, enabling swift adaptation to multi-label new classes. As shown in Fig.~\ref{Fig.mAP_base}~(b), our method demonstrates powerful learning capabilities for new classes with few parameters. Finally, we conduct extensive experiments to demonstrate that P2L-CA surpasses SOTA methods across both MLCIL and CIL benchmarks, notably achieving an improvement of up to \textbf{9.9\%} in final accuracy on the challenging COCO dataset.

The main contributions of this paper are summarized as follows:
\begin{itemize}
\setlength{\leftmargini}{0pt}

\item We propose P2L-CA, a parameter-efficient fine-tuning framework for MLCIL that operates without memory buffers and requires minimal trainable parameters.

\item We propose a P2L module that utilizes class-specific prompts to decouple multi-label features, while incorporating linguistic priors to enforce robust semantic-visual alignment.

\item We design a CA module to resolve domain discrepancies, thereby boosting the plasticity and adaptation efficiency of the model in downstream tasks

\item Extensive experiments demonstrate that P2L-CA outperforms SOTA methods across both standard and challenging MLCIL benchmarks, while also exhibiting robust generalization on the CIL task.
\end{itemize}

\section{Related Work}
\label{sec:rela}


\subsection{Single-label incremental learning.} 
At present, mainstream academic research on single-label incremental learning can be broadly categorized into the following two types:

(1) \textbf{Rehearsal-Based Methods}.
These approaches mitigate catastrophic forgetting by explicitly storing a subset of samples or class prototypes from previous tasks. For instance, ER~\cite{ERbase} and DER++~\cite{der+} maintain a memory buffer and replay stored instances during training on new tasks, often combined with distillation losses to reinforce prior knowledge. iCaRL~\cite{ICARL} and its variants~\cite{lucir,podnet,tao2020topology} further improve this paradigm by selecting representative exemplars via herding and designing more effective distillation objectives. Methods such as BIC~\cite{BIC} and IL2M~\cite{il2m} employ bias correction modules to alleviate distribution shifts between old and new classes. More recent approaches, including DyTox, dynamically expand network modules (e.g., task tokens), while DER~\cite{der} combines rehearsal with classifier expansion to better isolate representations across tasks. Nevertheless, these methods inherently face scalability limitations, as model size and memory requirements grow with the number of tasks.

(2) \textbf{Rehearsal-Free Methods}.
Compared with rehearsal-based approaches, rehearsal-free continual learning represents a stricter setting, where no samples, prototypes, or class statistics from previous tasks are stored. Early rehearsal-free methods mainly rely on regularization to preserve old knowledge. For example, EWC, oEWC, and SI~\cite{EWC,SI,oewc} constrain critical parameters based on their estimated importance (e.g., via the Fisher Information Matrix), while LwF~\cite{LWF,lwm}  and D-teacher~\cite{cil_csvt5} distill soft targets from the previous model to mitigate forgetting. However, although these methods comply with strict rehearsal-free constraints, they suffer from structural limitations: approximation errors accumulate as the number of tasks increases; their performance deteriorates in more challenging CIL settings; and they struggle to balance knowledge retention with the acquisition of new information, leading to a stability–plasticity trade-off. 

Recent studies have introduced parameter-efficient fine-tuning (PEFT) modules on top of frozen pre-trained backbones, providing a more powerful and scalable solution for rehearsal-free continual learning. Prompt-based methods, such as L2P~\cite{l2p} and DualPrompt~\cite{dualprompt}, learn a small set of insertable prompts to adapt to new tasks while keeping the backbone frozen, thereby significantly enhancing stability without requiring the storage of any task samples. Adapter- and LoRA-based approaches~\cite{cada,liang2024inflora,li2025dynamic} further decouple shared representations from task-specific parameters, mitigating query-matching inaccuracies in prompt methods and improving the stability–plasticity trade-off. Recently proposed hybrid methods, such as HiDe-Prompt~\cite{wang2023hierarchical} and SLCA~\cite{zhang2023slca}, enhance performance by incorporating class prototypes or class centers. However, these approaches implicitly store class-level statistics and therefore cannot be considered strictly rehearsal-free, as they resemble lightweight prototype-level rehearsal.

\subsection{Incremental object detection/segmentation.} 

Incremental object detection (IOD) and semantic segmentation (ISS) apply incremental learning to detection and segmentation tasks. In IOD, early work~\cite{iod2017} introduced knowledge distillation (KD) into Faster R-CNN’s output, laying the groundwork for the field. Subsequent studies~\cite{iod2022,IOD2021,dec_csvt4,det_csvt3} enhanced detection performance by incorporating KD into intermediate feature maps and region proposal networks or using exemplar replay (ER) to retain samples. Meanwhile, ER and KD have been successfully applied to Transformer-based frameworks, such as DETR~\cite{IOD2023}, broadening IOD’s methodological diversity. In ISS, approaches are generally categorized as regularization-based or replay-based. Regularization-based methods, such as PLOP~\cite{PLOP}, use KD strategies to constrain latent feature representations and prevent forgetting. Replay-based methods retain a small set of exemplars or pseudo-labels to represent earlier classes, effectively mitigating catastrophic forgetting~\cite{iss-2021er}. While the above IOD and CSS approaches effectively address MLCIL tasks when additional bounding boxes and per-pixel annotations are available, obtaining such detailed annotations is often costly. Therefore, advancing research on MLCIL using only image-level annotations holds considerable value and significance.

\subsection{Multi-label online incremental Learning}
Multi-label online incremental learning (MLOIL)~\cite{PRS_mloil} confronts severe challenges stemming from pervasive missing labels, which induce false negatives, and uncontrollable long-tailed class imbalance. To tackle these issues, PRS~\cite{PRS_mloil} employs a partitioning strategy to balance head and tail classes within the replay buffer, while OCDM~\cite{OCDM} optimizes class distribution in memory. Addressing the missing label problem, AGCN~\cite{AGCN} utilizes Graph Convolutional Networks to model dynamic label relationships. Crucially, AGCN offers a rehearsal-free variant, establishing it as a vital baseline for this study. More recently, CUTER~\cite{wang2025cut} leverages the localization ability of pretrained models, assessed via Laplacian spectral properties, to extract and replay label-specific regions.

However, most of these approaches address distribution shifts by caching representative samples or regions. Since this reliance on replay buffers violates the strict rehearsal-free constraint, they are excluded from our comparison.

\subsection{Multi-label class incremental Learning}
Multi-Label Class-Incremental Learning~\cite{KRT} (MLCIL) has advanced rapidly, with researchers proposing strategies to address challenges such as label absence, information dilution, and positive–negative imbalance. KRT~\cite{KRT} restores missing supervision via pseudo-labeling and preserves historical knowledge through an attention mechanism. APPLE~\cite{ML-SK} enhances replay-based methods by using Cluster Sampling for diverse memory buffering and a Decoder to mitigate feature dilution. As attention shifts toward prediction reliability, CSC~\cite{CSC2024} introduces a confidence self-calibration mechanism that models label dependencies with a GCN and suppresses overconfidence through max-entropy regularization. Subsequent works broaden these perspectives: HCP~\cite{HCP2025} manages the boundaries of past, present, and future knowledge via feature purification and prospective knowledge mining, while RebLL~\cite{rebll2025} addresses severe positive–negative imbalance through asymmetric knowledge distillation and online relabeling.

Despite their progress, most existing approaches remain model-based, relying on full-parameter backbone updates. This results in high computational and training costs and decreasing efficiency as incremental stages accumulate. To overcome these limitations, we propose a PEFT paradigm for MLCIL. Our rehearsal-free approach updates only a small set of trainable parameters, enabling efficient IL while substantially reducing computational overhead.

\begin{figure*}[htb]
\centering
\includegraphics[width=1\textwidth]{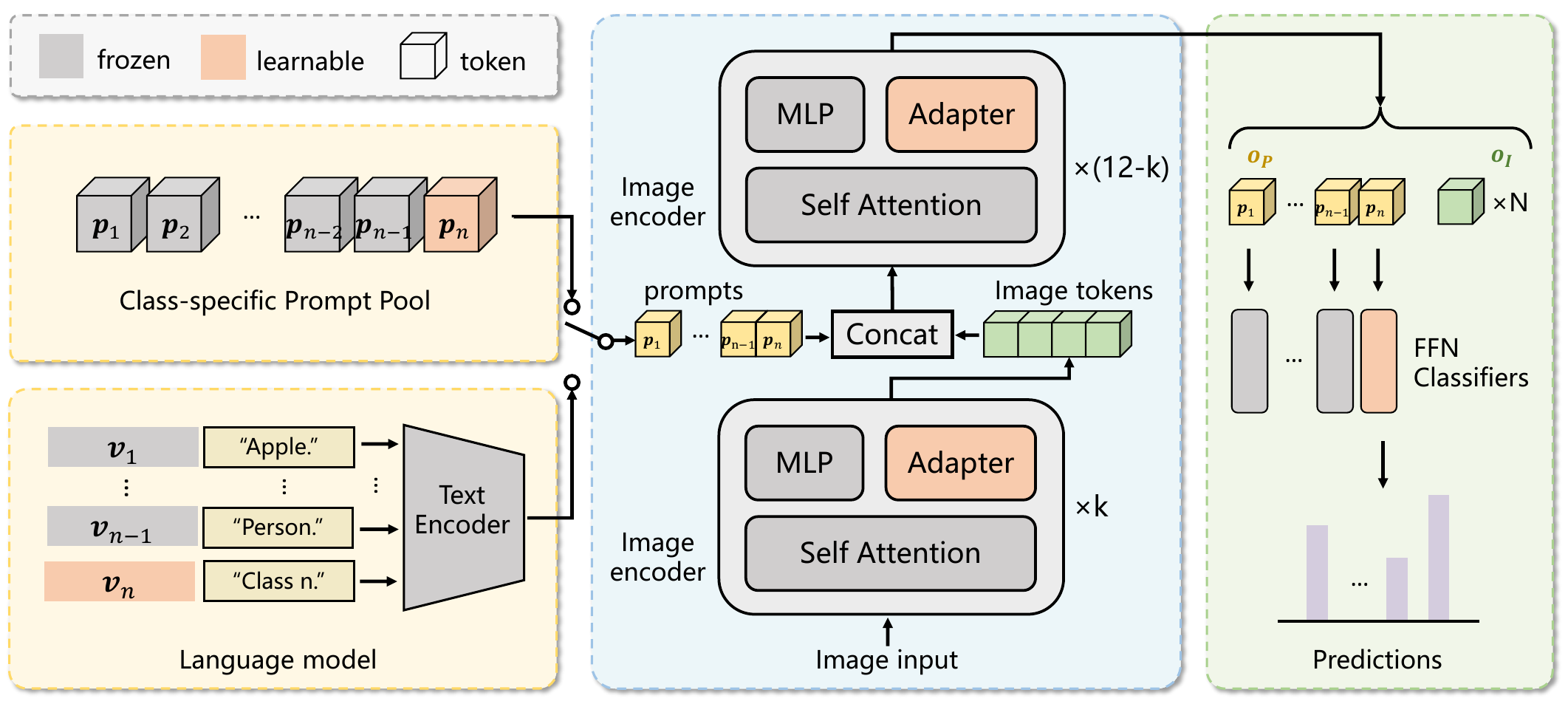}
\caption{The Framework of our proposed P2L-CA.
The image tokens are obtained from the first $k$ layers of the image encoder. Then they are concatenated with the prompt corresponding to each category and passed jointly to the subsequent encoder layers for self-attention computation. The prompts can be derived from a class-specific prompt pool with random initialization or a semantic feature initialized by a language model. The prediction is obtained by feeding the prompts part of the encoder output into the FFN classifiers.} 
\label{Fig.framework}
\end{figure*}

\section{Methodology}
\subsection{Problem Formulation}
Suppose we have a sequence of T training tasks ${D_1,D_2,\cdots,D_T}$, where $D_t=\{x_i^t, y_i^t\}$ is the $t$-th task datasets, with $x_i^t$ as the $i$-th training image and $y_i^t$ as its corresponding label of task $t$. The label space of each task is denoted by $C_t$, and we have $C_r\cap C_q=\emptyset$ when $r\neq q$, indicating that there are no overlapping classes between different tasks. The unified model is required to learn incrementally from the stream of training tasks while evaluating on a test set $E_t$ that contains all learned classes $C_{1-t}=C_1\cup C_2\cup\cdots\cup C_t$. In this paper, we focus on the exemplar-free setting of MLCIL, where no old task samples are stored.

\subsection{Overall Framework}

Fig.~\ref{Fig.framework} illustrates the framework of our proposed P2L-CA, which consists of two modules: P2L and CA. The P2L module comprises class-specific prompts and independent Feed Forward Network(FFN) classifiers, while the CA module embeds adapters into the layers of the image encoder to fine-tune its features. When an image is input, it is first processed by the first $k$ layers of the image encoder to obtain image tokens, which are then concatenated with $n$ class-specific prompts. The concatenated features are then fed into the remaining layers of the image encoder, resulting in the final output features ${o}$, which include the class-specific prompt part $\bm{o}_{P}$ and the image token part $\bm{o}_{I}$. To perform multi-label classification, we pass each token from the output $\bm{o}_{P}$ to its corresponding FFN classifier to determine whether that class is present in the image. Moreover, we can also use a text encoder to extract the features of each class name and replace the class-specific prompts with them for concatenation with the image tokens. Our P2L-CA+ method attempts to use CLIP’s text Transformer to enhance the model’s MLCIL capability.

\subsection{Prompt to Label module}

\noindent Prompt2Label framework comprises class-specific prompts and category-related FFN classifiers. Moreover, a language model can be employed to replace the class-specific prompts and further enhance the performance of MLCIL. We will elaborate on these three components in detail.

\noindent\textbf{Class-Specific Prompt Learning:} 
In the approach of Prompt2Label, we introduce using Class-Specific prompts $\bm{p}_i \in \mathbb{R}^{d}$, which are learnable parameters with the same dimension $d$ as the image token. When the model has learned $n$ classes, we have a prompt pool $\bm{P}={\{\bm{p}_i\}}^{n}_{i=1}$. When incorporating a new class, a prompt $\bm{p}_{n+1}$ is initialized and added to $\bm{P}$.
At the incremental stage $t$, the model is trained on the $D_t$ with the label set $C_t$. Those old classes prompts $\bm{P}_{old}=\{\bm{p}_i\}_{i \notin C_t}$ are frozen for preventing catastrophic forgetting, while only the prompts $\bm{P}_{new}=\{\bm{p}_i\}_{i \in C_t}$ corresponding to stage $t$ categories are learnable. 

For the input image, it is firstly patchified as $\bm{I} \in \mathbb{R}^{N\times d}$, where $N$ denotes the sequence length, and fed into the Image Encoder $\phi(\cdot)$. Each prompt in $\bm{P}$ is inserted into the $k$-th layer($k=6$ in this paper) of the Image Encoder and concatenated with image tokens. The total model is implemented as:

\begin{equation*}
  \begin{aligned}
    &x = \bm{I} \\
    &x = \phi^{l}(x), \,\, 1 \leq l \leq k,   \\
    &x^{\prime} = [\bm{p}_1, \bm{p}_2, \cdots, \bm{p}_i, x], \\
    &x^{\prime} = \phi^{l}(x^{\prime})), \,\, k+1 \leq l \leq 12\\
  \end{aligned}
\end{equation*}

where $\phi^{l}(\cdot)$ represents the $l$ layer of the image encoder, and the concatenation is denoted as $[\cdot]$. $x \in \mathbb{R}^{N\times d}$ is the image feature without prompts and $x^{\prime} \in \mathbb{R}^{(n+N)\times d}$ with prompt tokens. (For simplicity, we neglect the parameters of the CA module $\theta_c$). Finally, the output of the model $x^{\prime}$ contains two parts: $\bm{o}_{P} \in \mathbb{R}^{n \times d}$ corresponding to the prompt of each class and image patch tokens $\bm{o}_{I} \in \mathbb{R}^{N \times d}$. Only the $\bm{o}_{P}$ is fed into classification heads.\\

\noindent\textbf{Feed Forward Network Classifiers:} To establish a correspondence between class labels and class-specific prompts in $\bm{P}$, we only use the $\bm{o}_{P}$, which correspond to prompts, as the input of classifiers and use the independent FFN for classification. The FFN of class $i$ contains a weight $w_i \in \mathbb{R}^{d}$ and a bias $b_i$. Specifically, we have a total of $n$ independent FFNs as the classifiers of $n$ classes.
$$ p(y_i|x) = w_i^T\bm{o}_{P}^i + b_i, i \in [1, n] $$
where $ p(y_i|x) $ is the prediction probability of the $i$-th class. During the incremental stage $t$, assume that the new categories in stage $t$ are $C_t$, only the parameters of $\{w_i, b_i\}, i\in C_t$ are trainable, while others are frozen to prevent catastrophic forgetting.

\noindent\textbf{Language model:} 
To further enhance our model, we can leverage the language model $g(\cdot)$ to extract semantic information from the category name and leverage its semantic relationships in the feature space to improve the model’s multi-label classification performance. 
$$ \bm{P} = g([\bm{V}, \bm{E}]) $$

We can obtain the corresponding prompt pool $\bm{P}$ by applying $g(\cdot)$ to each category name.
Specifically, we can concatenate the word embeddings $\bm{E}=\{\bm{e}_1,\bm{e}_2,\dots,\bm{e}_n\}, \bm{e}_i \in \mathbf{R}^{d}$ and the random-initialized learnable context vectors $\bm{V}=\{\bm{v}_1,\bm{v}_2,\dots,\bm{v}_n\}, \bm{v}_i \in \mathbf{R}^{L_v \times d}$. And then feed them into $g(\cdot)$, obtaining the output $ \bm{P} \in \mathbb{R}^{n \times d} $, which represents the semantic feature of each category. 
The obtained $\bm{P}$ can act as the prompt pool and be concatenated with the image token as in the previous section.
In this way, the prompt initialized with semantic information can boost the model’s multi-label incremental learning ability. We use the CLIP pre-trained text transformer as the language model $g(\cdot)$ for this work.

\begin{figure}[h]
\centering
\includegraphics[width=0.6\linewidth]{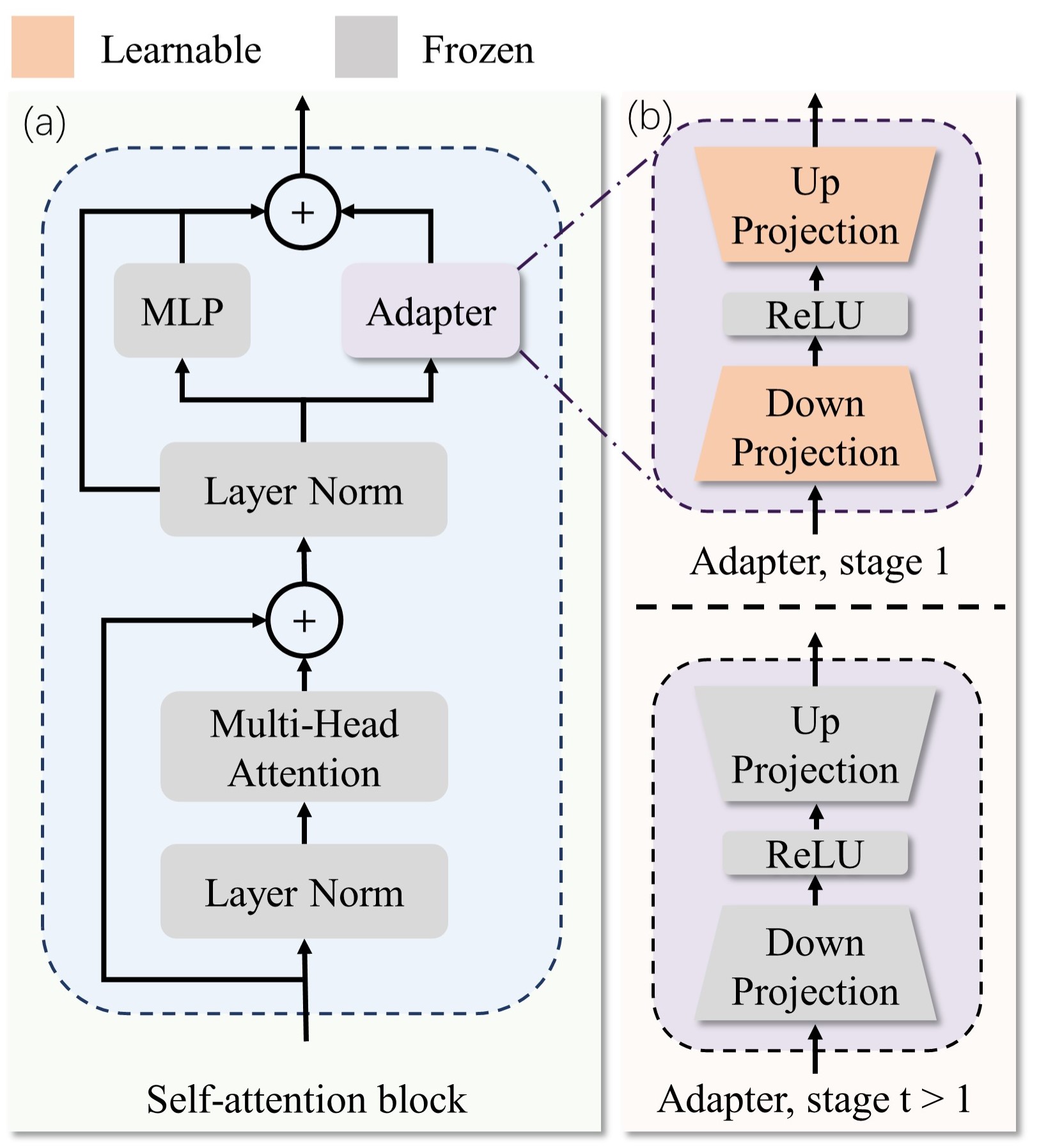}
\caption{(a) The illustration of the Continual-Adapter module in the Self-Attention Block. (b) Illustration of the adapter structure and the novel training paradigm at various stages.}
\label{Fig.adapter}
\end{figure}

\subsection{Continual-Adapter Module}
\noindent\textbf{Adapter Formulate:}
Adapter is a bottleneck module that consists of a down-projection layer $\rm Linear_{d\rightarrow d^\prime}$ to reduce the feature dimension, a non-linear activation function, and an up-projection layer $\rm Linear_{d^\prime \rightarrow d}$. We equip the original MLP structure in ViT with the adapter. Given the input of the adapter as $ x\in \mathbb{R}^{N\times d}$, the output of it is formatted as: 
\begin{equation*}
  \begin{aligned}
    y_{a} & = \rm{Adapter}(\it{x_a}) \quad where \, x_a, y_a \in \mathbb{R}^{\it{N\times d}}\\
    y_{a} &= {\rm{Linear}_{d^\prime \rightarrow d}}({\rm ReLU}({\rm Linear_{d\rightarrow d^\prime}}(x_a)))\\
  \end{aligned}
\end{equation*}
where output $y_{a}$ has the same size of input $x_a \in \mathbb{R}^{N\times d}$.

\noindent\textbf{Continual-Adapter:} We modify the original Self-Attention Block(SAB) of ViT, which consists of LayerNorm, multi-head self-attention, and the MLP layer. As shown in Fig.~\ref{Fig.adapter}, we insert the adapter module into the self-attention block parallel to the MLP layer. Assuming the $l$-th layer of the self-attention block is embedded, the modified SAB can be implemented as follows:
\begin{equation*}
  \begin{aligned}
    &x_o^l = \rm{MHSA}^{\it{l}}(\bm{LN}(\it{x^{l-1}})) + {x^{l-1}} \\
    &y_o^l = \rm{MLP}^{\it{l}}(\it{x}_o^l )+ {x}_o^l \\
    &y_a^l = \rm{Adapter}^{\it{l}}(\it{x}_o^l) \\
    &x^{l} = \rm{Add}(\it{y}_o^l,\,\,{y}_a^l)\\
  \end{aligned}   
\end{equation*}
where $x^{l-1}, x_o^l \in \mathbb{R}^{N\times d}$ is the input and output of the MHSA module, ${y}_o^l$, ${y}_a^l$ and ${x^{l}}$ are the output of MLP, Adapter and modified SAB.

To enhance the transfer learning ability for new domains, we embed multiple adapters into the Image Encoder $\phi(;\varphi_{I})$, and we term this set of adapters the Continual-Adapter module. Given that feature representations in the deeper layers exhibit a stronger correlation with downstream tasks, we strategically insert adapters exclusively into these layers, thus minimizing the introduction of extra parameters. The Image Encoder $\phi(;\varphi_{I})$ with the embedded CA module can be expressed as follows:

\begin{equation*}
  \begin{aligned}
    &x^0 = \bm{X}_p \\
    &x^l = \phi(x^{l-1};\varphi^{l}_{I}), \,\,\,\,\, {l} \in \{1,\cdots,m-1\} \\
    &x^l = \phi(x^{l-1};\varphi^l_{I},\,\theta_c^{l}), \,\, l \in \{m,\cdots,L\} \\
      \\
  \end{aligned}
\end{equation*}
where  $\bm{X}_p$ and $x^{l}$ denote the input image tokens and the output of the $l$-th layer, respectively. In this case, the adapters are embedded only in the layers after the $m$-th one, and $\theta_c$ represents the parameters of the CA module.\\  
\textbf{Continual-Adapter Learning Paradigm}: The proposed training paradigm is supposed to facilitate fast adaptation to new multi-label classes without any forgetting, as illustrated in Fig.~\ref{Fig.adapter}. It consists of two stages.\\
\textbf{First stage:} When $t=1$, the parameters of Continual-Adapter module $\theta_c$ are optimized to reduce the domain gap between datasets. (Image Encoder $\phi(;\varphi_{I})$ is frozen).  \\
\textbf{Continual stage:} When $t>1$, only the prompts and FFNs corresponding to the classes in the current stage are trained to quickly adapt to new classes. (The parameter of $\theta_c$ and $\varphi_{I}$ are frozen.)

\subsection{Optimization Objective}

Our model is only trained on $\mathbf{L}_{ASL}$: asymmetric loss~\cite{asl2020} as classification loss:
\begin{equation}
\label{eq6}
 L_{ASL}=\frac{1}{n} \sum_{j=1}^{n} \left\{
\begin{aligned}
 & (1-p_j)^{\gamma+}log(p_j), & y_j=1, \\
 & p_j^{\gamma-}log(1-p_j), & y_j=0,
\end{aligned}
\right.
\end{equation}

The proposed P2L-CA method does not involve any additional distillation loss or auxiliary loss, and there is no need for manual parameter tuning. This demonstrates the robustness and generality of our approach.

\begin{table*}[t]
\centering
\caption{Experimental results on the COCO dataset with mAP(\%) as the metric. The Impro. indicates the mAP gap compared to the P2L-CA method. The best scores are highlighted in \textbf{bold}, while the sub-optimal methods are marked with \underline{underline}. For the JOINT benchmark, results before parentheses correspond to our ViT backbone, and results in parentheses to the CNN backbone. For all experiments with CNN backbones, we report the accuracies as originally presented in~\cite{CSC2024,HCP2025}.
}
\renewcommand\arraystretch{1.5}
\small
\setlength\tabcolsep{4pt}
\begin{tabular}{l|c|c|c|cccccccc}
\toprule
\multirow{3}{*}{\textbf{Method}} & \multirow{3}{*}{\textbf{Backbone}} & \multirow{3}{*}{\textbf{PEFT}} & \multirow{3}{*}{\textbf{\begin{tabular}[c]{@{}c@{}}Buffer\\ size\end{tabular}}} & \multicolumn{8}{c}{\textbf{MS-COCO}} \\ \cline{5-12} 
 & & & & \multicolumn{4}{c|}{\textbf{B0-C10}} & \multicolumn{4}{c}{\textbf{B40-C10}} \\ \cline{5-12} 
 & & & & Last mAP & \multicolumn{1}{l|}{Impro.} & Avg. mAP & \multicolumn{1}{l|}{Impro.} & Last mAP & \multicolumn{1}{l|}{Impro.} & Avg. mAP & \multicolumn{1}{l}{Impro.} \\ \midrule
Joint & \multirow{2}{*}{-} & \multirow{2}{*}{-} & \multirow{2}{*}{-} & 83.1 (81.8) & \multicolumn{1}{c|}{-} & - & \multicolumn{1}{c|}{-} & 83.1 (81.8)  & \multicolumn{1}{c|}{-} & - & \multicolumn{1}{c}{-} \\
Fine-Tuning & & &  & 16.9 & \multicolumn{1}{c|}{-} & 38.3 & \multicolumn{1}{c|}{-} & 17.0 & \multicolumn{1}{c|}{-} & 35.1 & \multicolumn{1}{c}{-} \\
\bottomrule

ER~\cite{ERbase} & CNN & $\times$ & \multirow{4}{*}{20/class} & 47.2 & \multicolumn{1}{c|}{$\downarrow27.4$} & 60.3 & \multicolumn{1}{c|}{$\downarrow20.0$} & 61.6 & \multicolumn{1}{c|}{$\downarrow17.0$} & 68.9 & $\downarrow12.5$ \\
PODNet~\cite{podnet} & CNN & $\times$ &  & 58.8 & \multicolumn{1}{c|}{$\downarrow15.8$} & 70.0 & \multicolumn{1}{c|}{$\downarrow10.3$} & 64.2 & \multicolumn{1}{c|}{$\downarrow14.4$} & 71.0 & $\downarrow10.4$ \\
DER++~\cite{der+} & CNN & $\times$&  & 63.1 & \multicolumn{1}{c|}{$\downarrow11.5$} & 72.7 & \multicolumn{1}{c|}{$\downarrow7.6$} & 66.3 & \multicolumn{1}{c|}{$\downarrow12.3$} & 73.6 & $\downarrow7.8$ \\
KRT-R~\cite{KRT} & CNN & $\times$&  & 70.2 & \multicolumn{1}{c|}{$\downarrow4.4$} & 76.5 & \multicolumn{1}{c|}{$\downarrow3.8$} & 75.2 & \multicolumn{1}{c|}{$\downarrow3.4$} & 78.3 & $\downarrow3.1$\\ \midrule
oEWC~\cite{oewc} & CNN &  $\times$& \multirow{10}{*}{0/class} & 24.3 & \multicolumn{1}{c|}{$\downarrow50.3$} & 46.9 & \multicolumn{1}{c|}{$\downarrow33.4$} & 27.3 & \multicolumn{1}{c|}{$\downarrow51.3$} & 44.8 & $\downarrow36.6$ \\
LWF~\cite{LWF} & CNN &  $\times$&  & 28.9 & \multicolumn{1}{c|}{$\downarrow45.7$} & 47.9 & \multicolumn{1}{c|}{$\downarrow32.4$} & 29.9 & \multicolumn{1}{c|}{$\downarrow48.7$} & 48.6 & $\downarrow32.8$ \\
L2P~\cite{l2p} & ViT & $\checkmark$ &  & 67.7 & \multicolumn{1}{c|}{$\downarrow6.9$} & 73.3 & \multicolumn{1}{c|}{$\downarrow7.0$} & 70.4 & \multicolumn{1}{c|}{$\downarrow8.2$} & 73.1 & $\downarrow8.3$ \\
DualPrompt~\cite{dualprompt} & ViT & $\checkmark$ &  & 69.4 & \multicolumn{1}{c|}{$\downarrow5.2$} & 74.7 & \multicolumn{1}{c|}{$\downarrow5.6$} & 71.9 & \multicolumn{1}{c|}{$\downarrow6.7$} & 74.5 & $\downarrow6.9$ \\
CODA~\cite{coda-prompt} & ViT & $\checkmark$ &  & 69.8 & \multicolumn{1}{c|}{$\downarrow4.8$} & 75.8 & \multicolumn{1}{c|}{$\downarrow4.5$} & 72.4 & \multicolumn{1}{c|}{$\downarrow6.2$} & 75.3 & $\downarrow6.1$ \\
AGCN~\cite{AGCN} & CNN &  $\times$&  & 61.4 & \multicolumn{1}{c|}{$\downarrow13.2$} & 72.4 & \multicolumn{1}{c|}{$\downarrow7.9$} & 69.1 & \multicolumn{1}{c|}{$\downarrow9.5$} & 73.9 & $\downarrow7.5$ \\ 
KRT~\cite{KRT} & CNN &  $\times$&  & 65.9 & \multicolumn{1}{c|}{$\downarrow8.7$} & 74.6 & \multicolumn{1}{c|}{$\downarrow5.7$} & 74.0 & \multicolumn{1}{c|}{$\downarrow4.6$} & 77.8 & $\downarrow3.6$ \\ 
KRT*~\cite{KRT} & ViT &  $\times$&  & 66.4 & \multicolumn{1}{c|}{$\downarrow8.2$} & 74.9 & \multicolumn{1}{c|}{$\downarrow5.4$} & 73.6 & \multicolumn{1}{c|}{$\downarrow5.0$} & 77.2 & $\downarrow4.2$ \\
HCP~\cite{HCP2025} & CNN  &  $\times$&  & 71.2 & \multicolumn{1}{c|}{$\downarrow3.4$} & 77.9 & \multicolumn{1}{c|}{$\downarrow2.4$} & 75.3 & \multicolumn{1}{c|}{$\downarrow3.3$} & 78.9 & $\downarrow2.5$ \\ 
CSC~\cite{CSC2024} & CNN &  $\times$&  & 72.8 & \multicolumn{1}{c|}{$\downarrow1.8$} & 78.0 & \multicolumn{1}{c|}{$\downarrow2.3$} & 75.0 & \multicolumn{1}{c|}{$\downarrow3.6$} & 78.2 & $\downarrow3.2$ \\ 
 \rowcolor{gray!20} \textbf{P2L-CA} (Ours) & ViT & $\checkmark$ &  & \underline{74.6} & \multicolumn{1}{c|}{$\downarrow0.0$} & \underline{80.3} & \multicolumn{1}{c|}{\textbf{$\downarrow0.0$}} & \underline{78.6} & \multicolumn{1}{c|}{$\downarrow0.0$} & \underline{81.4} & \textbf{$\downarrow0.0$} \\
 \rowcolor{gray!40} \textbf{P2L-CA+} (Ours) & ViT & $\checkmark$ &  & \textbf{75.7} & \multicolumn{1}{c|}{$\uparrow1.1$} & \textbf{81.9} & \multicolumn{1}{c|}{\textbf{$\uparrow1.6$}} & \textbf{81.4} & \multicolumn{1}{c|}{$\uparrow2.8$} & \textbf{84.1} & \textbf{$\uparrow2.7$} \\ 
\bottomrule
\end{tabular}
\label{tab:coco}
\end{table*}

\section{Experiment} 
\subsection{Experimental Settings}

\noindent\textbf{Datasets}. To comprehensively evaluate the effectiveness of the proposed method, we conduct experiments on both multi-label and single-label CIL benchmarks. Following mainstream studies~\cite{KRT,CSC2024}, two widely used datasets in the multi-label recognition domain, MS COCO~\cite{coco2014} and PASCAL VOC~\cite{voc2007}, are adopted. In addition, we include CIFAR-100~\cite{cifar} to validate the generalizability of our method in conventional single-label CIL settings.

\textbf{MS COCO}~\cite{coco2014} is a large-scale and challenging benchmark containing 80 object categories. We use its official split, training on 82,081 images and validating on 40,137 images. The complex scenes and frequent object co-occurrences in COCO provide a rigorous testbed for evaluating model generalization under diverse multi-label conditions.

\textbf{PASCAL VOC}~\cite{voc2007} is a classic benchmark comprising 20 object categories. We follow the standard protocol, training on 5,011 images and evaluating on 4,952 images.

\textbf{CIFAR-100}~\cite{cifar} contains 60,000 color images across 100 classes, with 500 training and 100 test images per class. It serves as a standard benchmark for evaluating single-label CIL performance.

Together, these datasets cover diverse scales and label configurations, enabling a comprehensive and fair evaluation of our method across both multi-label and single-label incremental learning scenarios.

\noindent\textbf{Benchmark.} This study conducts a series of experiments on two standard benchmarks: Multi-Label Incremental Learning and Class-Incremental Learning. The incremental learning scenarios are uniformly defined using the Bx-Cy format. Herein, $x$ specifies the number of classes trained in the initial (base) task, and $y$ denotes the number of new classes introduced in each subsequent incremental stage. Following the completion of each incremental stage, the model's comprehensive performance is evaluated on the set of all classes encountered thus far, encompassing both the base classes and all previously learned incremental classes.

For the MLCIL experiments, our protocol strictly adheres to the established specifications in KRT~\cite{KRT}. Consistent with the methodologies in~\cite{KRT,CSC2024}, classes in all experiments are introduced according to the lexicographical order of their names. To thoroughly assess the model's robustness under varying levels of difficulty, our experiments encompass both standard protocols following traditional setups~\cite{CSC2024} and more challenging protocols based on recent work~\cite{rebll2025}. The standard protocols include the B10-C2 and B4-C4 settings on PASCAL VOC, as well as the B40-C10 and B0-C10 settings on MS-COCO. The more challenging protocols introduce a greater number of incremental stages or fewer initial classes, as seen in the B5-C3 and B4-C2 settings on PASCAL VOC, and the B20-C4 and B0-C5 settings on MS-COCO, aiming to provide a more rigorous evaluation benchmark.

For the CIL task, this study adopts an experimental setup consistent with advanced rehearsal-free methods~\cite{l2p,coda-prompt,cada}. Specifically, the experiments partition the CIFAR100 dataset into 10 independent and sequential learning stages (B0-C10).

\noindent\textbf{Evaluation Metric.} Following prior incremental learning studies~\cite{iod2022,KRT,lucir}, we employ several key metrics to assess the effectiveness of the proposed method comprehensively. For the primary MLCIL tasks, performance is measured using the Mean Average Precision (mAP), which evaluates the model’s accuracy across all learned categories in each incremental session. We report both the Average mAP (mean over all sessions) and the Final mAP (“Last. Acc”), corresponding to the performance in the final session. To further characterize the model’s ability to balance precision and recall, we also report the Per-class F1 (CF1) and Overall F1 (OF1) scores.

For the CIL tasks, following the evaluation protocol in~\cite{l2p}, we adopt the Final Accuracy (“Last. Acc”) and the Average Forgetting Measure (“Forgetting”). The former reflects the overall classification accuracy after completing all tasks, while the latter quantifies the extent of knowledge degradation on previous tasks as new ones are learned.

\begin{table*}[t!]
\centering
\renewcommand\arraystretch{1.3}
\small
\caption{Experimental results on the VOC dataset with mAP(\%) as the metric. The best scores are highlighted in \textbf{bold}, while the sub-optimal methods are marked with \underline{underline}. For the JOINT benchmark, results before parentheses correspond to our ViT backbone, and results in parentheses to the CNN
backbone. The reported accuracies for the compared methods are sourced directly from the papers~\cite{CSC2024,HCP2025}}
\setlength\tabcolsep{4pt}
\setlength{\belowcaptionskip}{-0.1cm}
\begin{tabular}{l|c|c|cc|cc|cc|cc}
\hline
\multicolumn{1}{c|}{\multirow{3}{*}{\textbf{Method}}} & \multirow{3}{*}{\textbf{PEFT}} & \multirow{3}{*}{\textbf{\begin{tabular}[c]{@{}c@{}}Buffer\\ size\end{tabular}}} & \multicolumn{8}{c}{\textbf{PASCAL VOC}} \\ \cline{4-11} 
\multicolumn{1}{c|}{} &  &  & \multicolumn{2}{c|}{\textbf{B0-C4}} & \multicolumn{2}{c|}{\textbf{B10-C2}} & \multicolumn{2}{c|}{\textbf{B5-C3}} & \multicolumn{2}{c}{\textbf{B4-C2}} \\ \cline{4-11} 
\multicolumn{1}{c|}{} &  &  & Last mAP & \multicolumn{1}{c|}{Avg. mAP} & Last mAP & \multicolumn{1}{c|}{Avg. mAP} & Last mAP & \multicolumn{1}{c|}{Avg. mAP} & Last mAP & Avg. mAP \\ \hline
Joint  &  \multirow{2}{*}{-} & \multirow{2}{*}{-} & 94.1(93.6) & \multicolumn{1}{c|}{-} & 94.1(93.6) & \multicolumn{1}{c|}{-} & 94.1(93.6) & \multicolumn{1}{c|}{-} & 94.1(93.6) & - \\ 
Fine-Tuning &  &  & 62.9 & \multicolumn{1}{c|}{ 82.1} & 43.0 & \multicolumn{1}{c|}{70.1} & 49.4 & \multicolumn{1}{c|}{74.4} & 37.0 & 60.4 \\ 
\midrule
ER~\cite{ERbase} & $\times$ & \multirow{3}{*}{2/class}  & 71.5 & \multicolumn{1}{c|}{86.1} & 68.6 & \multicolumn{1}{c|}{81.5} & 65.3 & \multicolumn{1}{c|}{79.0} & 57.9 & 73.7 \\
PODNet~\cite{podnet} & $\times$ &  & 76.6 & \multicolumn{1}{c|}{88.1} & 71.4 & \multicolumn{1}{c|}{81.2} & 70.3 & \multicolumn{1}{c|}{81.4} & 62.5 & 75.7 \\
DER++~\cite{der+} & $\times$ &  & 76.1 & \multicolumn{1}{c|}{87.9} & 70.6 & \multicolumn{1}{c|}{82.3} & 68.1 & \multicolumn{1}{c|}{78.0} & 61.6 & 77.0 \\
\midrule
L2P~\cite{l2p} & $\checkmark$ & \multirow{10}{*}{0} & 82.9 & \multicolumn{1}{c|}{89.7} & 82.2 & \multicolumn{1}{c|}{87.1} & 79.2 & \multicolumn{1}{c|}{86.9} & 70.1 & 81.6 \\
Dual-Prompt~\cite{dualprompt} & $\checkmark$ &  & 82.9 & \multicolumn{1}{c|}{89.6} & 81.8 & \multicolumn{1}{c|}{87.8} & 79.7 & \multicolumn{1}{c|}{87.5} & 70.6 & 82.2 \\
CODA~\cite{coda-prompt} & $\checkmark$ &  & 83.7 & \multicolumn{1}{c|}{89.9} & 82.0 & \multicolumn{1}{c|}{89.7} & 80.4 & \multicolumn{1}{c|}{88.0} & 71.3 & 82.5 \\
AGCN~\cite{AGCN} & $\times$ &  & 73.4 & \multicolumn{1}{c|}{84.3} & 65.1 & \multicolumn{1}{c|}{79.4} & 69.2 & \multicolumn{1}{c|}{80.9} & 53.4 & 72.1 \\ 
KRT*~\cite{KRT}  & $\times$ &  &  81.8& \multicolumn{1}{c|}{89.3} & 80.9 & \multicolumn{1}{c|}{88.5} &78.5& \multicolumn{1}{c|}{88.3} &68.7& 82.3 \\ 
HCP~\cite{HCP2025}  & $\times$ &  &  \underline{87.9} & \multicolumn{1}{c|}{92.9} & 81.9 & \multicolumn{1}{c|}{90.1} &-  & \multicolumn{1}{c|}{-} & - & - \\ 
RebLL~\cite{rebll2025}  & $\times$ &  & -& \multicolumn{1}{c|}{-} & - & \multicolumn{1}{c|}{-} & 79.4 & \multicolumn{1}{c|}{87.7} &73.1& 84.6 \\ 
CSC~\cite{CSC2024}   & $\times$ &  & 85.1 & \multicolumn{1}{c|}{90.4} & 83.8 & \multicolumn{1}{c|}{89.0} & 82.1 & \multicolumn{1}{c|}{88.0} & 74.1 & 83.3 \\ 
\rowcolor{gray!20} P2L-CA(Ours) & $\checkmark$ &  &87.1 & \multicolumn{1}{c|}{\underline{93.2}} & \underline{86.4} & \multicolumn{1}{c|}{\underline{92.2}} & \underline{84.4} & \multicolumn{1}{c|}{\underline{90.4}} & \underline{78.7} & \underline{87.3} \\
  \rowcolor{gray!40} P2L-CA+(Ours) & $\checkmark$ &  & \textbf{88.1} & \multicolumn{1}{c|}{\textbf{93.9}} & \textbf{86.9} & \multicolumn{1}{c|}{\textbf{92.9}} & \textbf{85.8} & \multicolumn{1}{c|}{\textbf{91.3}} & \textbf{79.9} & \textbf{88.2} \\ \hline
\end{tabular}
\label{tab:voc}
\end{table*}

\noindent\textbf{Comparison Methods.} To comprehensively evaluate the proposed P2L-CA method (and its enhanced variant P2L-CA+), we conduct extensive comparisons with representative and SOTA approaches on both MLCIL and CIL tasks. Following prior studies, the compared methods are grouped into rehearsal-based and rehearsal-free categories.

For the MLCIL task, the rehearsal-based methods include classic CIL approaches adapted for the multi-label setting, such as ER~\cite{ER}, iCaRL~\cite{ICARL}, TPCIL~\cite{tao2020topology}, DER++~\cite{der+}, and PODNet~\cite{podnet}. The rehearsal-free group comprises regularization-based approaches (oEWC~\cite{oewc}, LwF~\cite{LWF}) and recent prompt-based methods (L2P~\cite{l2p}, DualPrompt~\cite{dualprompt}, CODA-Prompt~\cite{coda-prompt}) that leverage pre-trained ViT-B backbones. Notably, for the model-based MLCIL methods such as KRT~\cite{KRT}, HCP~\cite{HCP2025}, CSC~\cite{CSC2024}, and RebLL~\cite{rebll2025}, while they typically employ replay buffers, they also offer rehearsal-free variants. To ensure a fair comparison with our proposed approach, we specifically adopt the rehearsal-free versions of these methods in our experiments. For consistency, we also re-implement KRT using the ViT-B/16 backbone.

For the CIL task, we follow a similar grouping: rehearsal-based methods (ER~\cite{ER}, BiC~\cite{BIC}, GDumb~\cite{GDumb}, C$o^2$L~\cite{co2l}) and rehearsal-free ones, including EWC~\cite{EWC}, LwF~\cite{LWF}, L2P~\cite{l2p}, DualPrompt~\cite{dualprompt}, CODA-Prompt~\cite{coda-prompt}, and the adapter-based method C-ADA~\cite{cada}. We exclude parameter-efficient fine-tuning methods~\cite{wang2023hierarchical,zhang2023slca} that store or replay class centers, as they violate the strict “rehearsal-free” setting.

\noindent\textbf{Implementation Details.} We implement our method in PyTorch using two NVIDIA RTX 4090 GPUs. The ViT-B/16 model pretrained on ImageNet-21K~\cite{ImageNet} is adopted as our image encoder $\phi(\cdot)$. It is worth noting that our reproduction experiments demonstrate that directly applying KRT to a ViT-S/16 backbone fails to yield any performance improvement over its original TResNet-M implementation, confirming that the gains reported in our work stem from the proposed methodology rather than the backbone architecture. For hyperparameter configuration, the insertion depth of the Class-Specific Prompts is set to $k = 6$, and the number of Continual-Adapters is set to $9$. The model is trained with the Adam optimizer using a batch size of 64. The initial learning rate is set to $4 \times 10^{-4}$ and follows a cosine-annealing decay schedule. Each incremental stage is trained for 20 epochs. We use the Asymmetric Loss (ASL)~\cite{asl2020} without additional distillation or auxiliary objectives.

\begin{table*}[t]
\centering
\renewcommand\arraystretch{1.3}
\small
\caption{Experimental results on the MS-COCO dataset under the challenge protocols. The best scores are highlighted in \textbf{bold}. The reported accuracies for the compared methods are sourced directly from the paper~\cite{rebll2025}.}
\label{tab:challenge}
\begin{tabular}{l|c|ccc|c|ccc|c}
\hline
\multirow{2}{*}{Method} & \multirow{2}{*}{Buffer} 
& \multicolumn{4}{c}{MS-COCO B20-C4} & \multicolumn{4}{c}{MS-COCO B0-C5} \\
\cline{3-10}
 & & \multicolumn{3}{c}{Last Acc} & Avg.Acc 
   & \multicolumn{3}{c}{Last Acc} & Avg.Acc \\
\cline{3-10} 
 & & mAP & CF1 & OF1 & mAP & mAP & CF1 & OF1 & mAP \\
\hline
Joint & \multirow{2}{*}{-} & 83.1(81.8) & 77.6(76.4) & 80.1(79.4) & - & 83.1(81.8) & 77.6(76.4) & 80.1(79.4) & - \\
Fine-Tuning &  & 19.4 & 10.9 & 13.4 & 36.5 & 22.5 & 15.0 & 23.6 & 48.1 \\ 
\midrule
ER~\cite{ER} & \multirow{4}{*}{5/class}  & 41.9 & 32.9 & 29.8 & 53.0 & 40.1 & 32.9 & 32.3 & 54.6 \\
PODNet~\cite{podnet} &  & 58.4 & 44.0 & 39.1 & 67.7 & 58.2 & 45.1 & 40.8 & 67.2 \\
DER++~\cite{der+} &  & 57.3 & 41.4 & 35.5 & 65.5 & 57.9 & 43.6 & 39.2 & 68.2 \\
RebLL~\cite{rebll2025} & & 62.8 & 53.3 & 53.0 & 71.2 & 65.5 & 56.1 & 54.0 & 72.0 \\ \midrule
LwF~\cite{LWF}  & \multirow{6}{*}{0} & 34.6 & 17.3 & 31.8 & 55.4 & 50.6 & 36.3 & 41.1 & 66.2 \\
AGCN~\cite{AGCN}&& 55.6 & 44.2 & 39.6 & 65.7 & 53.0 & 43.2 & 41.1 & 64.4 \\
KRT~\cite{KRT} & & 45.2 & 17.6 & 33.0 & 64.0 & 44.5 & 22.6 & 37.5 & 63.1 \\
CSC~\cite{CSC2024} &  & 60.6 & 44.5 & 43.0 & 69.8 & 63.4 & 50.7 & 50.1 & 71.1 \\
RebLL~\cite{rebll2025} &  & 60.1 & 51.3 & 49.2 & 69.2 & 63.5 & 53.5 & 51.9 & 71.7 \\
 \rowcolor{gray!20} P2L-CA (Ours) &  & \textbf{70.5} & \textbf{69.2} & \textbf{64.8} & \textbf{75.7} & \textbf{70.8} & \textbf{68.7} & \textbf{66.3} & \textbf{76.3} \\
\hline
\end{tabular}
\end{table*}

\subsection{Comparison Results on Standard Protocols}

\noindent\textbf{Results on COCO.} 
Table \ref{tab:coco} reports the quantitative results on MS-COCO, where \textbf{P2L-CA} consistently achieves SOTA performance.
\begin{itemize}
    \item \textbf{P2L-CA vs. Rehearsal-Free Methods:} P2L-CA establishes a new benchmark among exemplar-free methods. Under the B40-C10 protocol, it significantly outperforms the leading model-based method, HCP~\cite{HCP2025}, improving the Last mAP from 75.3\% to 78.6\%. This improvement validates that our class-level prompts capture fine-grained semantics more effectively than traditional approaches, thereby mitigating category confusion. Furthermore, attributed to our CA module, which bridges the domain gap between tasks, P2L-CA surpasses the PEFT-based CODA-Prompt~\cite{coda-prompt} by 6.2\% on B40-C10. Notably, even under identical ViT architectures, P2L-CA maintains a lead of 8.2\% over KRT on B0-C10, confirming the intrinsic superiority of our framework.
    \item \textbf{P2L-CA vs. Rehearsal-based Methods:} Remarkably, despite being rehearsal-free, P2L-CA outperforms the replay-based KRT-R (20/class) by 4.4\% in Last mAP and 3.8\% in Avg mAP on B0-C10. This demonstrates its capability to overcome catastrophic forgetting without the storage and privacy costs associated with memory buffers.
     \item \textbf{Effectiveness of Semantic Enhancement:} Finally, incorporating CLIP semantic priors (P2L-CA+) yields further performance gains. Under the B40-C10 protocol, the Last mAP improves by 2.8\%. This confirms that leveraging rich semantic features in complex incremental scenarios significantly optimizes the extraction of class-specific features, thereby enhancing the IL efficacy.
\end{itemize}

\noindent\textbf{Results on VOC:} We observe similar superior conclusions on the VOC dataset, as presented in Table~\ref{tab:voc}. Our proposed \textbf{P2L-CA} and \textbf{P2L-CA+} frameworks consistently achieve the highest scores, establishing new SOTA results for rehearsal-free MLCIL.

\begin{itemize} 
\item \textbf{P2L-CA vs. Rehearsal-Free Methods:} On the \textbf{B10-C2} benchmark, P2L-CA achieves a Last mAP of 86.4\% and an Avg. mAP of 92.2\%. This performance surpasses the sub-optimal non-exemplar method CSC by a substantial margin of 2.6\% in Last mAP and 3.2\% in Avg. mAP, validating the effectiveness of our class-level P2L module. 
\item \textbf{P2L-CA vs. Rehearsal-Based Methods:} Despite using a buffer size of 0, P2L-CA (Last mAP 86.4\% on B10-C2) robustly outperforms all exemplar-based methods (buffer size 2/class) by a significant margin. \end{itemize}

\subsection{Compare Results on Challenge Protocols:}

To further verify the robustness of our method in stricter scenarios, we evaluated P2L-CA on Challenge Protocols~\cite{rebll2025}, which involve longer incremental sequences or fewer initial categories. These include the high-stage protocols on VOC (B5-C3 and B4-C2) and the stringent settings on MS-COCO (B20-C4 and B0-C5). As shown in Table~\ref{tab:voc} and Table~\ref{tab:challenge},  compared methods suffer from severe catastrophic forgetting under these conditions. In contrast, our \textbf{P2L-CA} outperforms other methods by a large margin.

\begin{table}[t]
\renewcommand\arraystretch{1.2}
\centering
\setlength\tabcolsep{4pt}
\small
\caption{Performance comparison on CIFAR-100. The reported accuracies for the compared methods are sourced directly from the paper~\cite{dualprompt,cada}}
\label{tab:results}
\begin{tabular}{l|c|cc}
\toprule
\multirow{2}{*}{Method} & \multicolumn{3}{c}{CIFAR-100} \\
\cmidrule(lr){2-4}
                        & Buffer Size & Last. Acc ($\uparrow$)       & Forgetting ($\downarrow$)    \\
\midrule
ER~\cite{ERbase}             & \multirow{4}{*}{1000}        & 67.87$\pm$0.57      & 33.33$\pm$1.28      \\
BiC~\cite{BIC}           &      & 66.11$\pm$1.76      & 35.24$\pm$1.64      \\
GDumb~\cite{GDumb}          &          & 67.14$\pm$0.37      & -                \\
Co$^2$L~\cite{co2l}            &          & 72.15$\pm$1.32      & 28.55$\pm$1.56      \\ \midrule
ER~\cite{ERbase}             &  \multirow{4}{*}{5000}        & 82.53$\pm$0.17      & 16.46$\pm$0.25      \\
BiC~\cite{BIC}           &        & 81.42$\pm$0.85      & 17.31$\pm$1.02      \\
GDumb~\cite{GDumb}          &        & 81.67$\pm$0.02      & -                \\
Co$^2$L~\cite{co2l}            &          & 82.49$\pm$0.89      & 17.48$\pm$1.80      \\ \midrule
FT-seq                  &  \multirow{8}{*}{0}            & 33.61$\pm$0.85      & 86.87$\pm$0.20      \\
EWC~\cite{EWC}             &           & 47.01$\pm$0.29      & 33.27$\pm$1.17      \\
LwF~\cite{LWF}             &           & 60.69$\pm$0.63      & 27.77$\pm$2.17      \\
L2P~\cite{l2p}             &           & 83.86$\pm$0.28      & 7.35$\pm$0.38       \\
DualPrompt~\cite{dualprompt}             &           & 86.51$\pm$0.33      & 5.16$\pm$0.09       \\
CODA~\cite{coda-prompt}             &          & 86.94$\pm$0.63      & 4.04$\pm$0.09       \\
C-ADA~\cite{cada}      &   & 87.18$\pm$0.28   &  4.01$\pm$0.08 \\

\rowcolor{gray!20} P2L-CA  &          & 88.12$\pm$0.35      & 3.95$\pm$0.07       \\ \midrule

Joint             &    -       & 90.85$\pm$0.12      & -                \\
\bottomrule
\end{tabular}
\end{table}

\begin{itemize} 

\item \textbf{VOC Challenge Tasks (B5-C3 and B4-C2):} Our method demonstrates even more pronounced advantages in the challenging long-sequence protocols compared to standard settings. Specifically, under the B4-C2 protocol, \textbf{CA-P2L} outperforms the previous SOTA by 4.6\% in Last mAP, a significantly larger margin than the 2.6\% gain observed in the B10-C2 protocol. Furthermore, the text-enhanced variant, \textbf{CA-P2L+}, yields an additional improvement of 1.2\%.

\item \textbf{COCO Challenge Tasks (B20-C4 and B0-C5):} The difficulty is further escalated on the MS-COCO dataset. For instance, under the B20-C4 setting, LwF achieves a Last mAP of only 34.6\%, underscoring the rigorous nature of this benchmark. In contrast, \textbf{P2L-CA} achieves the highest performance with a Last mAP of 70.5\% and an Average mAP of 75.7\%. This surpasses the runner-up method, CSC (60.6\%), by a remarkable margin of nearly 10\%, significantly wider than the typical 2-3\% gap seen in standard protocols. Similarly, \textbf{P2L-CA} maintains its leadership in the B0-C5 protocol, reaching a Last mAP of 70.8\%.
\end{itemize}

These results strongly demonstrate that the P2L-CA framework possesses exceptional anti-forgetting ability and sustained learning capability when handling extreme challenge scenarios. This validates the efficiency and robustness of our proposed parameter tuning strategy and module design in overcoming the key challenges inherent in MLCIL tasks.

\subsection{Compare Results on Single-label Protocols:}
To validate the generalizability of our method in conventional single-label CIL settings, we compare \textbf{P2L-CA} against established baselines on CIFAR-100, as reported in Table~\ref{tab:results}. Strictly adhering to the rehearsal-free setting, we exclude parameter-efficient fine-tuning methods that rely on storing class centers for replay~\cite{wang2023hierarchical,zhang2023slca}. Among rehearsal-free approaches, our method achieves a new state-of-the-art Last Accuracy of 88.12\%, outperforming the recent adapter-based C-ADA and prompt-based CODA-Prompt. Specifically, it surpasses the previous best result by 0.94\% while further minimizing the forgetting rate to a new low of 3.95\%. More notably, \textbf{P2L-CA} significantly outperforms traditional rehearsal-based methods, even those utilizing a large buffer size of 5,000 (e.g., ER: 82.53\%, Co$^2$L: 82.49\%), demonstrating its exceptional effectiveness in mitigating catastrophic forgetting without requiring any historical data storage.

\subsection{Ablation Study}
To thoroughly understand the contribution of each component in our P2L-CA framework, we conduct extensive ablation experiments on the COCO B40-C10 and B0-C10 benchmarks. The analysis focuses on three aspects: (1) the effectiveness of the CA and P2L modules, (2) the necessity of freezing strategies during incremental learning, and (3) the design choices for class-specific prompts.\\

\begin{table}[]
\centering
\caption{Results of ablating components of the proposed P2L-CA method on COCO B40C10 and COCO B0C10 benchmark.}
\small
\setlength{\abovecaptionskip}{-0.1cm} 
\setlength{\belowcaptionskip}{-0.1cm}
\renewcommand\arraystretch{1.15}
\setlength\tabcolsep{5pt}

\begin{tabular}{cc|cc|cc}
\hline
\multicolumn{2}{c|}{\textbf{Method}} & \multicolumn{2}{c}{\textbf{COCO B40C10}} & \multicolumn{2}{c}{\textbf{COCO B0C10}} \\ 
\cline{1-6}
\textbf{CA} & \textbf{P2L} & \textbf{Last mAP} & \textbf{Avg mAP} &\textbf{Last mAP} & \textbf{Avg mAP}\\ \hline

 &  & 69.7 & 71.9  &67.8 &71.1 \\
\checkmark &  & 76.3 & 79.7  &73.0 &79.4 \\
 & \checkmark & 76.8 & 79.8 &73.5 & 79.5\\
\rowcolor{gray!20} \checkmark & \checkmark & \textbf{78.6} & \textbf{81.4} & \textbf{74.6} & \textbf{80.3}\\ 
\hline

\end{tabular}
\label{ab}
\end{table}

\noindent\textbf{The Effectiveness of Each Component.} 
Starting from a ViT-B/16 baseline trained with asymmetric loss, we progressively introduced the CA and P2L modules. The results are shown in Table~\ref{ab}.
(1). Adding the CA module alone improves the last mAP\textbf{ from 69.7\% to 76.3\%} on COCO B40-C10 and \textbf{from 67.8\% to 73.0\%} on COCO B0-C10, demonstrating its ability to enhance feature transferability from the pre-trained domain to the downstream dataset. This confirms that the CA effectively mitigates domain discrepancy and improves generalization to new tasks.

(2). Introducing the P2L module alone yields a comparable improvement, reaching\textbf{ 76.8\%} last mAP on B40-C10 and \textbf{73.5\%} on B0-C10. The gains mainly come from its capability to disentangle class-level representations through class-specific prompts, which reduces feature confusion among co-occurring categories.

(3). When both modules are jointly applied, the model achieves \textbf{the best} performance, with \textbf{78.6\%/81.4\%} (Last/Avg mAP) on B40-C10 and \textbf{74.6\%/80.3\%} on B0-C10. These results verify that CA and P2L complement each other: CA focuses on improving transferability and stability, while P2L enhances class discrimination and plasticity.\\

\noindent\textbf{Importance of Freezing Strategy.} In our framework, the CA module is trained exclusively during the initial stage and frozen in subsequent incremental steps. Similarly, the P2L module optimizes only the prompts corresponding to the current classes, while prompts for previously learned categories remain fixed. To validate the necessity of this design, we examine the impact of parameter freezing in Table~\ref{ab2}. When the CA module remains trainable rather than frozen, the performance suffers a significant degradation, with the last mAP dropping to \textbf{68.2\%} (under the B40-C10 setting). Furthermore, unfreezing the prompts for previous classes leads to an additional decline, reducing the last mAP to \textbf{66.9\%}. These results underscore that freezing both the adapters and historical prompts is critical for mitigating catastrophic forgetting. This frozen architecture effectively stabilizes the feature space, preventing the corruption of established knowledge while allowing new prompts and classifiers to adapt efficiently to novel categories. \\

\noindent\textbf{Further Analysis of Class-Specific Prompts.} As illustrated in the t-SNE visualization in Fig.~\ref{Fig.mAP_base}~(a), our method effectively mitigates semantic confusion, yielding well-separated clusters for distinct categories. To further investigate whether the Class-Specific Prompts capture independent and discriminative knowledge, we introduced an auxiliary orthogonal loss~\cite{coda-prompt} to regularize the prompt learning process. This loss is designed to maximize the orthogonality of the prompts $P$ in the feature space, theoretically minimizing interference between new and frozen prompts. However, as reported in Table~\ref{ab2}, the incorporation of the orthogonal loss yielded no performance improvement. This suggests that our proposed method is inherently robust against inter-class interference, as the learned prompts naturally acquire sufficiently discriminative representations without requiring explicit orthogonality constraints.\\

\begin{table}[]
\centering
\caption{Analysis of Freezing Strategies and Prompt Designs in P2L-CA on COCO B40C10 and B0C10 benchmarks.}
\setlength{\abovecaptionskip}{-0.1cm} 
\setlength{\belowcaptionskip}{-0.1cm}
\renewcommand\arraystretch{1.3}
\small
\setlength\tabcolsep{3pt}

\begin{tabular}{l|cc|cc}
\hline
{\textbf{Method}}  & \multicolumn{2}{c|}{\textbf{COCO B40C10}} & \multicolumn{2}{c}{\textbf{COCO B0C10}} \\
 & \textbf{Last mAP} & \textbf{Avg mAP} & \textbf{Last mAP} & \textbf{Avg mAP} \\ 
\hline\hline

\multicolumn{5}{l}{\textit{(a) Ablation on Freezing Params}} \\ \hline
CA Unfrozen       & 68.2 & 75.1 & 67.5 & 74.0 \\
Prompt Unfrozen   & 66.9 & 73.6 & 66.2 & 72.8 \\ 
\hline\hline

\multicolumn{5}{l}{\textit{(b) More Analysis of Class-Specific Prompts}} \\ \hline
P2L-CA + ortho.   & 78.6 & 81.2 & 74.3 & 80.1 \\ 
P2L-CA            & \textbf{78.6} & \textbf{81.4} & \textbf{74.6} & \textbf{80.3} \\
\hline

\end{tabular}
\label{ab2}
\end{table}

\begin{figure}[t] 
\setlength{\abovecaptionskip}{0.1 cm} 
\setlength{\belowcaptionskip}{-0.6 cm}
\centering    
\includegraphics[width=0.83\linewidth]{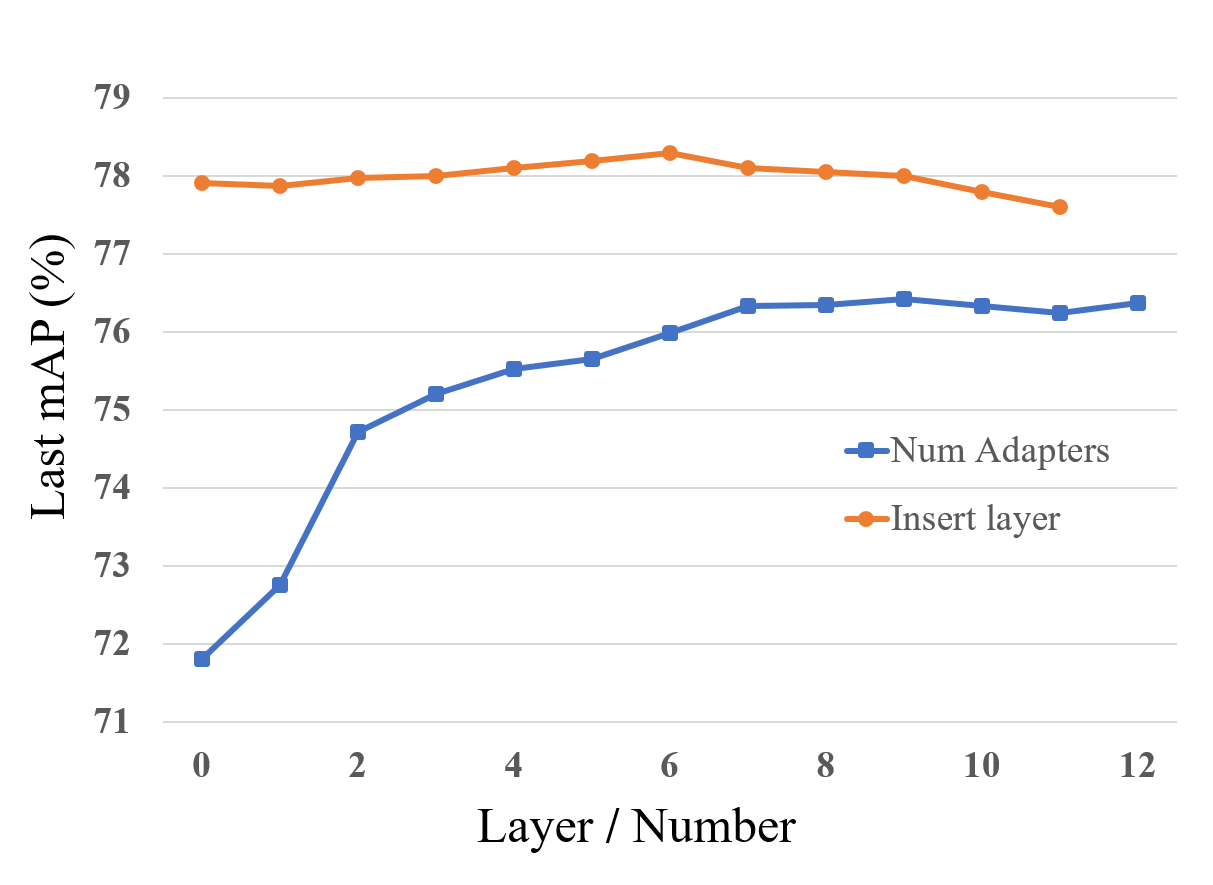}
\caption{Influence of the layers of equipped adapters and inserted prompts}
\label{fig.ablation}     
\end{figure}

\noindent \textbf{Effect of Insertion Depth and Adapter Number.} As illustrated in Fig.~\ref{fig.ablation}, we further investigated two architectural hyperparameters.
First, varying the insertion layer of prompts reveals that injecting prompts at the \textbf{6th} Transformer block achieves the highest accuracy (78.4\% mAP). \textbf{Earlier} layers underperform due to \textbf{insufficient semantic abstraction}, while \textbf{later} insertions reduce the contextual interaction between prompts and visual tokens.
Second, we conducted experiments to investigate the effect of the number of adapters in the Continual-Adapter (CA) module. As illustrated in Fig.~\ref{fig.ablation}, the network accuracy drops sharply when fewer than two adapters are used, indicating that an adequate number of adapters is essential for effectively adapting the pre-trained model to incremental learning tasks. As the number of adapters increases, the performance continues to improve and reaches its peak at nine adapters; beyond this point, the accuracy saturates, and no further gains are observed. Considering the trade-off between parameter efficiency and model performance, we adopt \textbf{nine} adapters~(layer 4-12) in the CA module to achieve effective adaptation to new datasets.



\begin{figure}[h]
\centering
 \setlength{\abovecaptionskip}{0.1cm} 
\setlength{\belowcaptionskip}{-0.3cm} 
\includegraphics[width=0.9\linewidth]{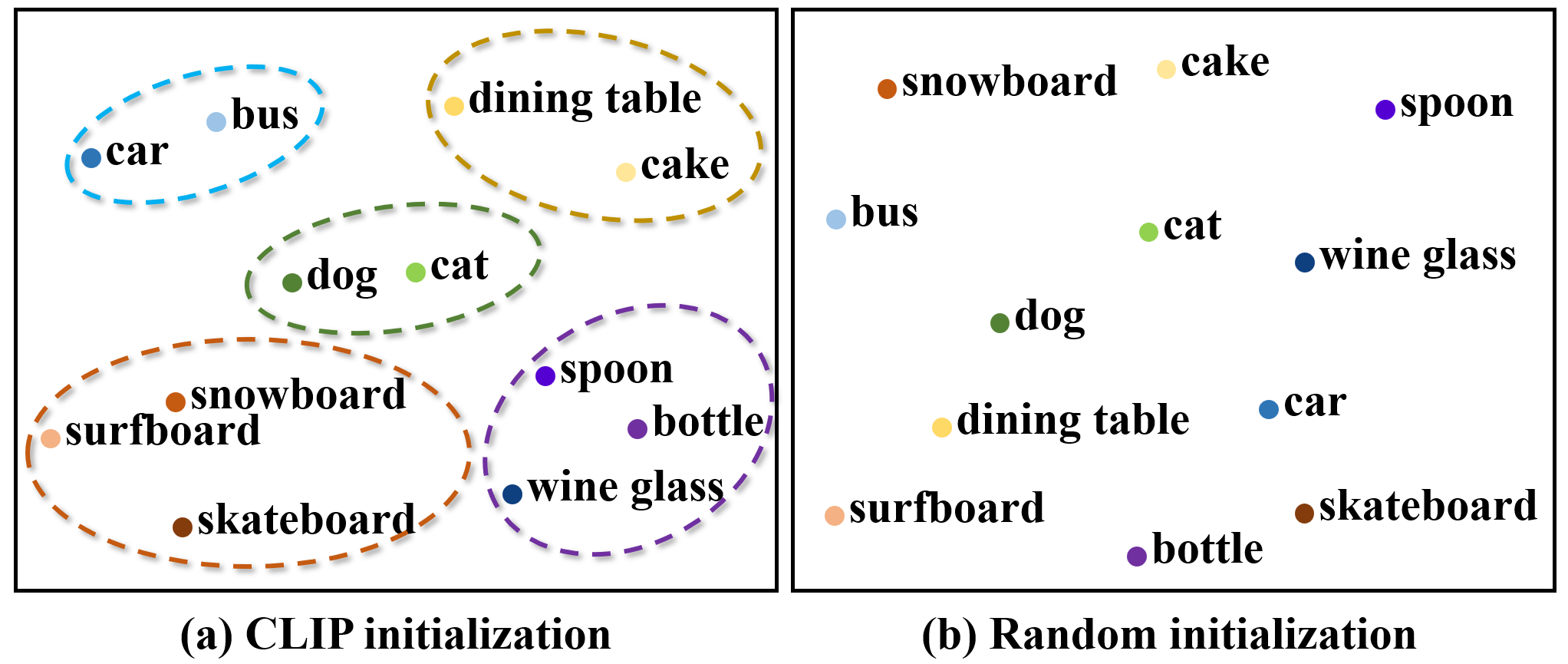}
\caption{t-SNE visualization of prompt tokens in the feature space}
\label{Fig.vis_prompt_tokens}
\end{figure}

\noindent\textbf{Visualization of Text Embedding.} 
The t-SNE visualization of the learned prompts further demonstrates that class-specific tokens occupy well-separated regions in the feature space. Prompts initialized using language models (CLIP) implicitly capture the semantic relationships between categories. As shown in Fig.~\ref{Fig.vis_prompt_tokens} (a), semantically similar or related classes are positioned closer together in the feature space, such as dog and cat, car and bus, or dining table and cake, whereas randomly initialized prompts (Fig.~\ref{Fig.vis_prompt_tokens} (b)) do not exhibit this property. These observations indicate that language-initialized prompts form semantically meaningful clusters, where visually or conceptually related categories are closer. This property enables P2L-CA+ to better understand and model inter-class relationships in the multi-label class-incremental learning problem, particularly for frequently co-occurring classes, and aligns with the quantitative results, demonstrating that P2L-CA+ effectively learns representations that are both semantically coherent and incrementally stable.

\section{Conclusion}
This paper presents P2L-CA, a parameter-efficient framework designed for MLCIL under strict rehearsal-free settings. By integrating a Prompt-to-Label (P2L) mechanism with a Continuous Adapter (CA) module, our approach effectively addresses catastrophic forgetting without relying on exemplar buffers or full-parameter fine-tuning. Our modular design offers distinct insights for the field. First, the P2L module demonstrates that class-specific prompts can effectively disentangle multi-label representations in complex scenes. This provides a viable strategy for mitigating feature confusion through explicit class-level anchors rather than implicit global optimization. Second, the CA module employs a specialized training paradigm to bridge the domain discrepancy between pre-trained models and incremental tasks. Its freezing strategy successfully balances model stability and plasticity, enabling efficient adaptation with minimal parameter overhead. Furthermore, our experiments with P2L-CA+ highlight the benefit of integrating linguistic priors from pre-trained language models (e.g., CLIP). Initializing prompts with semantic embeddings facilitates robust semantic-visual alignment. In conclusion, P2L-CA achieves state-of-the-art performance on benchmarks such as MS-COCO and PASCAL VOC. Its privacy-preserving nature, requiring no historical data storage, combined with high computational efficiency, establishes it as a practical paradigm for real-world applications.

\noindent\textbf{Limitation.} First, the Continuous Adapter module employs a "first-stage adaptation followed by freezing" strategy; this design relies on the static assumption that the domain distribution of subsequent tasks remains consistent, thereby limiting flexibility in handling continuous dynamic domain shifts. Second, certain classic baselines are based on CNN architectures; although reproduction experiments indicate that directly transferring these methods to ViT yields suboptimal results, and our method demonstrates a more significant relative performance improvement over the Joint upper bound compared to these baselines, this backbone heterogeneity nonetheless adds complexity to direct horizontal comparisons. Third, the current framework is confined to image-level classification tasks and has not yet been extended to fine-grained perception tasks such as object detection or segmentation, restricting its direct applicability in scenarios requiring precise spatial localization.

\noindent\textbf{Future work}. Future work will focus on three avenues: exploring adaptive dynamic tuning mechanisms to address continuous domain shifts, extending the Prompt-to-Label paradigm to fine-grained tasks such as object detection and segmentation, and leveraging large language models to exploit deeper semantic priors for open-world.


\ifCLASSOPTIONcaptionsoff
  \newpage
\fi

\bibliographystyle{IEEEtran}
\bibliography{ref}


\end{document}